# Features and Agreement

Sam Bayer and Mark Johnson*
Cognitive and Linguistic Sciences, Box 1978
Brown University
{bayer,mj}@cog.brown.edu

## Abstract

This paper compares the *consistency-based* account of agreement phenomena in 'unification-based' grammars with an *implication-based* account based on a simple feature extension to Lambek Categorial Grammar (LCG). We show that the LCG treatment accounts for constructions that have been recognized as problematic for 'unification-based' treatments.

## 1 Introduction

This paper contrasts the treatment of agreement phenomena in standard complex feature structure or 'unification-based' grammars such as HPSG (Pollard and Sag, 1994) with that of perhaps the simplest possible feature extension to Lambek Categorial Grammar (LCG) (Lambek, 1958). We identify a number of situations where the two accounts make different predictions, and find that generally the LCG account is superior. In the process we provide analyses for a number of constructions that have been recognized as problematic for 'unification-based' accounts of agreements (Zaenen and Karttunen, 1984; Pullum and Zwicky, 1986; Ingria, 1990). Our account builds on the analysis of coordination in applicative categorial grammar in Bayer (1994) and the treatment of Boolean connectives in LCG provided by Morrill (1992). Our analysis is similiar to that proposed by Mineur (1993), but differs both in its application and details.

The rest of the paper is structured as follows. The next section describes the version of LCG we use in this paper; for reasons of space we assume familiarity with the treatment of agreement in 'unification-based' grammars, see Shieber (1986) and Pollard and Sag (1994) for details. Then each of the following sections up to the conclusion discusses an important difference between the two approaches.

## 2 Features in Lambek Categorial Grammar

In LCG semantic interpretation and long distance dependencies are handled independently of the feature system, so agreement phenomena seem to be the major application of a feature system for LCG. Since only a finite number of feature distinctions need to be made in all the cases of agreement we know of, we posit only a very simple feature system here. Roughly speaking, features will be treated as atomic propositions (we have no need to separate them into attributes and values), and a simple category will be a Boolean combination of such atomic 'features' (since we have no reason to posit a recursive feature structures either). In fact we are agnostic as to whether more complex feature systems for LCG are linguistically justified; in any event Dorre et. al. (1994) show how a full attribute-value feature structure system having the properties described here can be incorporated into LCG.

Following the standard formulation of LCG, we regard the standard LCG connectives '/' and '\' as *directed implications*, so we construct our system so that $\alpha/\beta$ $\beta'$ can combine to form $\alpha$ if $\beta'$ is logically stronger than $\beta$.

Formally, we adopt Morrill's treatment (Morrill, 1992) of the (semantically impotent) Boolean connectives '∧' and '∨' (Morrill named these '⊓' and '⊔' respectively). Given a set of atomic features $\mathcal{F}$, we define the set of feature terms $\mathcal{T}$ and categories $\mathcal{C}$ as follows, where '/' and '\' are the standard LCG forward and backward implication operators.

$$\begin{array}{rcl} \mathcal{T} & ::= & \mathcal{F} + \mathcal{T}\wedge\mathcal{T} + \mathcal{T}\vee\mathcal{T} \\ \mathcal{C} & ::= & \mathcal{T} + \mathcal{C}/\mathcal{C} + \mathcal{C}\backslash\mathcal{C} \end{array}$$

In general, atomic categories in a standard categorial grammar will be replaced in our analyses with formulae drawn from $\mathcal{T}$. For example, the NP *Kim* might be assigned by the lexicon to the category $np \wedge sg \wedge 3$, the verb *sleeps* to the category $s\backslash np \wedge sg \wedge 3$, and the verb *slept* (which does not impose person or number features on its subject) to the category $s\backslash np$.

---

*We would like to thank Bob Carpenter, Pauline Jacobson, John Maxwell, Glynn Morrill and audiences at Brown University, the University of Pennsylvania and the Universität Stuttgart for helpful comments on this work. Naturally all errors remain our own.

To simplify the presentation of the proofs, we formulate our system in natural deduction terms, and specify the properties of the Boolean connectives using the single inference rule $P$, rather than providing separate rules for each connective.

$$\frac{\phi}{\psi}P \text{ where } \phi \vdash \psi \text{ in the propositional calculus.}^1$$

The rule $P$ allows us to replace any formula in $\mathcal{T}$ with a logically weaker one. For example, since *Kim* is assigned to the category $np \wedge sg \wedge 3$, then by rule $P$ it will belong to $np$ as well.

Finally, we assume the standard LCG introduction and elimination rules for the directed implication operators.

$$\frac{A/B \quad B}{A}/e \qquad \frac{B \quad A\backslash B}{A}\backslash e$$

$$\frac{[B]^n}{\vdots} \qquad \frac{[B]^n}{\vdots}$$
$$\frac{A}{A/B}/i^n \qquad \frac{A}{A\backslash B}\backslash i^n$$

For example, the following proof of the well-formedness of the sentence *Kim slept* can be derived using the rules just given and the lexical assignments described above.

$$\frac{\dfrac{\text{Kim}}{\dfrac{np \wedge sg \wedge 3}{np}P} \quad \dfrac{\text{slept}}{s\backslash np}}{s}\backslash e$$

This example brings out one of the fundamental differences between the standard treatment of agreement in 'unification-based' grammar and this treatment of agreement in LCG. In the 'unification-based' accounts agreement is generally a *symmetric* relationship between the agreeing constituents: both agreeing constituents impose constraints on a shared agreement value, and the construction is well-formed iff these constraints are *consistent*.

However, in the LCG treatment of agreement proposed here agreement is inherently *asymmetric*, in that an argument must logically imply, or be *subsumed* by, the antecedent of the predicate it combines with. Thus in the example above, the rule $P$ could be used to 'weaken' the argument from $np \wedge sg \wedge 3$ to $np$, but it would not allow $np$ (without agreement features) to be 'strengthened' to, say, $np \wedge sg \wedge 3$.

Abstracting from the details of the feature systems, we can characterize the 'unification-based' approach as one in which agreement is possible between two constituents with feature specifications $\phi$ and $\psi$ iff $\phi$ and $\psi$ are consistent, whereas the LCG approach requires that the argument $\phi$ implies the corresponding antecedent $\psi$ of the predicate (i.e., $\phi \models \psi$).

Interestingly, in cases where features are fully specified, these subsumption and consistency requirements are equivalent. More precisely, say that a formula $\phi$ from a feature constraint language *fixes* an atomic feature constraint $\chi$ iff $\phi \models \chi$ or $\phi \models \neg \chi$. For example, in single-valued feature systems $\langle person \rangle = 1$ and $\langle person \rangle = 3$ both fix $\langle person \rangle = 1$, $\langle person \rangle = 2$, $\langle person \rangle = 3$, etc., and in general all fully-specified agreement constraints fix the same set of formulae.

Now let $\phi$ and $\psi$ be two satisfiable formulae that fix the same set of atomic feature constraints. Then $\phi \wedge \psi$ is consistent iff $\phi \models \psi$. To see this, note that because $\phi$ and $\psi$ fix the same set of formulae, each condition holds iff $\phi$ and $\psi$ are elementarily equivalent (i.e., for each feature constraint $\chi$, $\phi \models \chi$ iff $\psi \models \chi$).

However, the role of *partial* agreement feature specifications in the two systems is very different. The following sections explore the empirical consequences of these two approaches. We focus on coordination phenomena because this is the one area of the grammar where underspecified agreement features seem to play a crucial linguistic role, and cannot be regarded merely as an abbreviatory device for a disjunction of fully-specified agreement values.

## 3 Coordination and agreement asymmetries

Interestingly, the analysis of coordination is the one place where most 'unification-based' accounts abandon the symmetric consistency-based treatment of agreement and adopt an asymmetric subsumption-based account. Working in the GPSG framework Sag et. al. (1985) proposed that the features on a conjunction must be the most specific category which subsumes each conjunct (called the *generalization* by Shieber (1992)). Shieber (1986) proposed a weaker condition, namely that the features on the conjunction must *subsume* the features on each conjunct, as expressed in the annotated phrase struc-ture rule below (Shieber, 1992).[2] In all of the examples we discuss below, the features associated with a conjunction is the generalization of the features associated with each of its conjuncts, so our conclusions are equally valid for both the generalization and subsumption accounts of coordination.

---

[1] Because conjunction and disjunction are the only connectives we permit, it does not matter whether we use the classical or intuitionistic propositional calculus here. In fact, if categories such as $np$ and $ap$ are 'decomposed' into the conjunctions of atomic features $+noun \wedge -verb$ and $+noun \wedge +verb$ respectively as in the Sag et. al. (1985) analysis discussed below, disjunction is not required in any of the LCG analyses below. However, Bayer (1994) argues that such a decomposition is not always plausible.

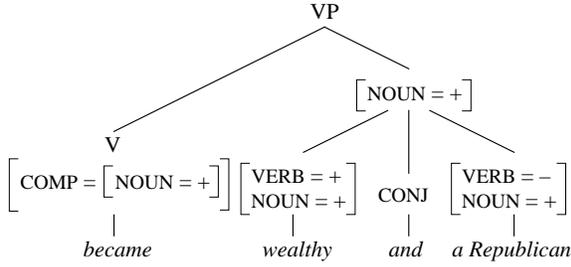

Figure 1: The feature structure subsumption analysis of (2b).

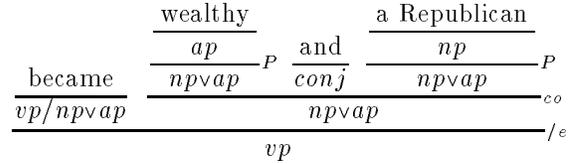

Figure 2: The LCG analysis of (2b).

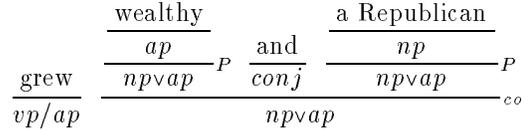

Figure 3: A blocked LCG analysis of the ungrammatical (3b).

(1) $X_0 \longrightarrow X_1 \; conj \; X_2$
where $X_0 \sqsubseteq X_1$ and $X_0 \sqsubseteq X_2$

Consider the sentences in (2). Decomposing the categories N(oun) and A(djective) into the Boolean-valued features $\{\langle noun \rangle = +, \langle verb \rangle = -\}$ and $\{\langle noun \rangle = +, \langle verb \rangle = +\}$ respectively, the fact that *became* can select for either an NP or an AP complement (2a) can be captured by analysing it as subcategorizing for a complement whose category is underspecified; i.e., its complement satisfies $\langle noun \rangle = +$, and no constraint is imposed on the *verb* feature.

(2) a. Kim [V became ] [AP wealthy ] / [NP a Republican ]

   b. Kim [VP [V became ] [AP wealthy ] and [NP a Republican ] ]

Now consider the coordination in (2b). Assuming that *became* selects the underspecified category $\langle noun \rangle = +$, the features associated with the coordination subsume the features associated with each coordinate, as required by rule (1), so (2b) has the well-formed structure shown in Figure 1.

On the other hand, a verb such as *grew* which selects solely AP complements (3a) requires that its complement satisfies $\langle noun \rangle = +, \langle verb \rangle = +$. Thus the features on the coordinate structure in (3b) must include $\langle verb \rangle = +$ and so do not subsume the $\langle verb \rangle = -$ feature on the NP complement, correctly predicting the ungrammaticality of (3b).

(3) a. Kim grew [AP wealthy]/*[NP a Republican]

   b. *Kim [VP [V grew ] [AP wealthy ] and [NP a Republican ] ]

Our LCG account analyses these constructions in a similar way. Because the LCG account of agreement has subsumption 'built in', the coordination rule merely requires identity of the conjunction and each of the conjuncts.

$$\frac{\vdots \quad \vdots}{A} co$$
$$A \quad conj \quad A$$

Condition: No undischarged assumptions in any conjunct.[3]

We provide an LCG derivation of (2b) in Figure 2. Roughly speaking, rule $P$ allows both the AP *wealthy* and the NP *a Republican* to 'weaken' to $np \vee ap$, so the conjunction satisfies the antecedent of the predicate *became*. (This weakening also takes place in non-coordination examples such as *Kim became wealthy*). On the other hand, (3b) is correctly predicted to be ill-formed because the strongest possible category for the coordination is $np \vee ap$, but this does not imply the 'stronger' $ap$ antecedent of *grew*, so the derivation in Figure 3 cannot proceed to form a $vp$.

Thus on these examples, the feature-based subsumption account and the LCG of complement coordination constructions impose similiar feature constraints; they both require that the predicate's feature specification of the complement subsumes the features of each of the arguments. In the feature-based account, this is because the features associated with a conjunction must subsume the features associated with each conjunct, while in the LCG account the features associated with the complement specification in a predicate must subsume those associated with the complement itself.

---

[2] Note that the LFG account of coordination provided by Kaplan and Maxwell (1988) differs significantly from both the generalization and the subsumption accounts of coordination just mentioned, and does not generate the incorrect predictions described below.

[3] This condition in effect makes conjunctions into islands. Morrill (1992) shows how such island constraints can be expressed using modal extensions to LCG.

Now consider the related construction in (4) involving conjoined predicates as well conjoined arguments. Similar constructions, and their relevance to the GPSG treatment of coordination, were first discussed by Jacobson (1987). In such cases, the feature-based subsumption account requires that the features associated with the predicate conjunction subsume those associated with each predicate conjunct. This is possible, as shown in Figure 4. Thus the feature structure subsumption account incorrectly predicts the well-formedness of (4).

(4) *Kim [ grew and remained ] [ wealthy and a Republican ].

Because the subsumption constraint in the LCG analysis is associated with the predicate-argument relationship (rather than the coordination construction, as in the feature-based subsumption account), an LCG analysis paralleling the one given in Figure 4 does not exist. By introducing and withdrawing a hypothetical *ap* constituent as shown in Figure 5 it is possible to conjoin *grew* and *remained*, but the resulting conjunction belongs to the category *vp/ap*, and cannot combine with the *wealthy and a Republican*, which belongs to the category *np*∨*ap*.

Informally, while rule $P$ allows the features associated with an *argument* to be *weakened*, together with the introduction and elimination rules it permits the *argument specifications of predicates* to be *strengthened* (c.f. the subproof showing that *remained* belongs to category *vp/ap* in Figure 5). As we remarked earlier, in LCG predicates are analysed as (directed) implicational formulae, and the argument features required by a predicate appear in the *antecedent* of such formulae. Since strengthening the antecedent of an implication weakens the implication as a whole, the combined effect of rule $P$ and the introduction and elimination rules is to permit the overall weakening of a category.

## 4 Consistency and agreement

Complex feature structure analyses of agreement require that certain combinations of feature constraints are *inconsistent* in order to correctly reflect agreement failure. For example, the agreement failure in *him runs* is reflected in the inconsistency of the constraints $\langle case \rangle = acc$ and $\langle case \rangle = nom$. In the LCG account presented above, the agreement failure in *him runs* is reflected by the failure of *acc* to imply *nom*, not by the inconsistency of the features *acc* and *nom*. Thus in LCG there is no principled reason not to assign a category an apparently contradictory feature specification such as $np \wedge nom \wedge acc$ (this might be a reasonable lexical category assignment for an NP such as *Kim*).

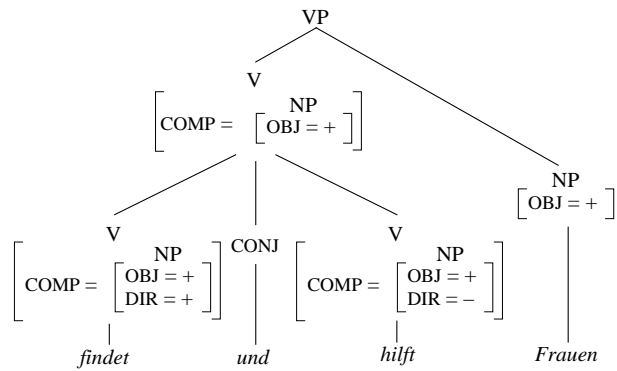

Figure 6: The feature structure subsumption analysis of (5c).

Consider the German examples in (5), cited by Pullum and Zwicky (1986) and Ingria (1990). These examples show that while the conjunction *findet und hilft* cannot take either a purely accusative (5a) or dative complement (5b), it can combine with the NP *Frauen* (5c), which can appear in both accusative and dative contexts.

(5)  a. * Er findet und hilft Männer
        he find-ACC and help-DAT men-ACC

     b. * Er findet und hilft Kindern
        he find-ACC and help-DAT children-DAT

     c.   Er findet und hilft Frauen
        he find-ACC and help-DAT women-ACC+DAT

Contrary to the claim by Ingria (1990), these examples can be accounted for straight-forwardly using the standard feature subsumption-based account of coordination. Now, this account presupposes the existence of appropriate underspecified categories (e.g., in the English example above it was crucial that major category labels were decomposed into the features *noun* and *verb*). Similarly, we decompose the four nominal cases in German into the 'subcase' features *obj* (abbreviating 'objective') and *dir* (for 'direct') as follows.

| Nominative | $\{\langle dir \rangle = +, \langle obj \rangle = -\}$ |
|---|---|
| Accusative | $\{\langle dir \rangle = +, \langle obj \rangle = +\}$ |
| Dative | $\{\langle dir \rangle = -, \langle obj \rangle = +\}$ |
| Genetive | $\{\langle dir \rangle = -, \langle obj \rangle = -\}$ |

By assigning the NPs *Männer* and *Kindern* the fully specified case features shown above, and *Frauen* the underspecified case feature $\langle obj \rangle = +$, both the feature structure generalization and subsumption accounts of coordination fail to generate the ungrammatical (5a) and (5b), and correctly accept (5c), as shown in Figure 6.

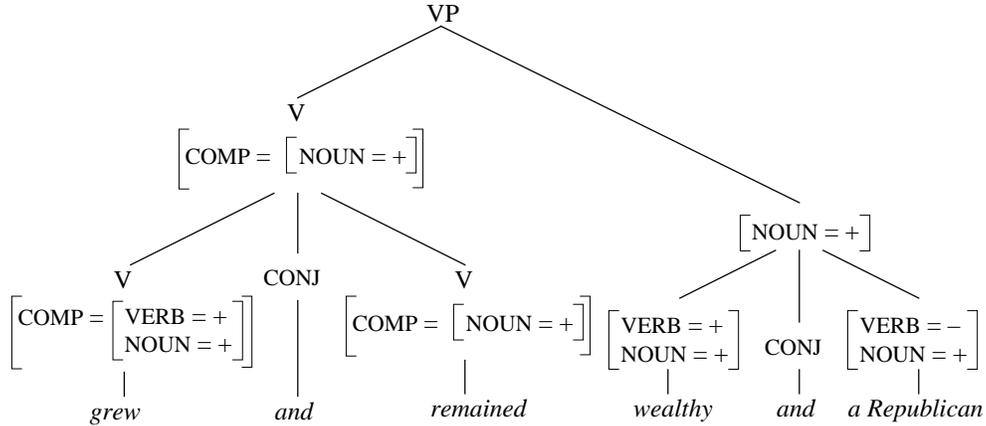

Figure 4: The feature structure subsumption analysis of the ungrammatical (4).

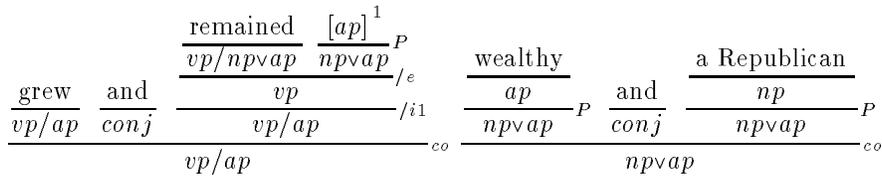

Figure 5: A blocked LCG analysis of the ungrammatical (4).

As in the previous example, the LCG approach does not require the case feature to be decomposed. However, as shown in Figure 7 it does assign the conjunction *findet und hilft* to the category $vp/np \wedge acc \wedge dat$; hence the analysis requires that *Frauen* be assigned to the 'inconsistent' category $np \wedge acc \wedge dat$. Such overspecified or 'inconsistent' features may seem ad hoc and unmotivated, but they arise naturally in the formal framework of Morrill's extended LCG.

In fact, they seem to be necessary to obtain a linguistically correct description of coordination in German. Consider the ungrammatical 'double coordination' example in (6). Both the feature structure generalization and subsumption accounts incorrectly predict it to be well-formed, as shown in Figure 8.

(6) * Er findet und hilft Männer und
     he find-ACC and help-DAT men-ACC and
     Kindern
     children-DAT

However, the LCG analysis systematically distinguishes between *Frauen*, which is assigned to the category $np \wedge acc \wedge dat$, and *Männer und Kindern*, which is assigned to the weaker category $np \wedge (acc \vee dat)$. Thus the LCG analysis correctly predicts (6) to be ungrammatical, as shown in Figure 9. The distinction between the categories $np \wedge acc \wedge dat$ and $np \wedge (acc \vee dat)$, and hence the existence of the apparently inconsistent categories, seems to be crucial to the ability to distinguish between the grammatical (5c) and the ungrammatical (6).

## 5 Conclusion

This paper has examined some of the differences between a standard complex feature-structure account of agreement, which is fundamentally organized around a notion of *consistency*, and an account in an extended version of LCG, in which agreement is fundamentally an asymmetric relationship. We have attempted to show that the LCG account of agreement correctly treats a number of cases of coordination which are problematic for the standard feature-based account. Although we have not shown this here, the LCG account extends straightforwardly to the cases of coordination and morphological neutralization discussed by Zaenen and Kartunen (1984), Pullum and Zwicky (1986) and Ingria (1990).

The nature of an appropriate feature system for LCG is still an open question. It is perhaps surprising that the simple feature system proposed here can handle such complex linguistic phenomena, but additional mechanisms might be required to treat other linguistic constructions. The standard account of adverbial modification in standard LCG, for instance, treats adverbs as functors. Because the verb heading an adverbial modified VP agrees in number with its subject, the same number features will have to appear in both the antecedent and consequent of the adverb. Using the LCG account described above it is necessary to treat adverbs as ambiguous, assigning them to the categories $(s\backslash np \wedge sg)\backslash(s\backslash np \wedge sg)$ and $(s\backslash np \wedge pl)\backslash(s\backslash np \wedge pl)$.

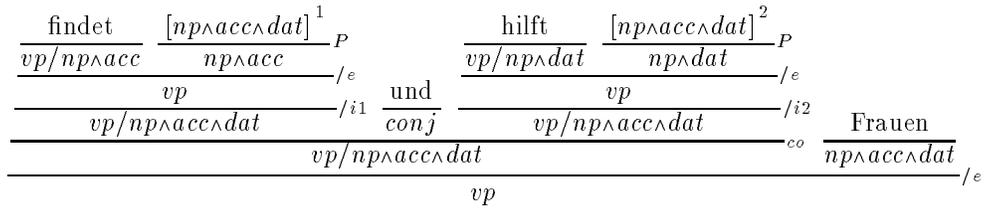

Figure 7: The LCG analysis of (5c)

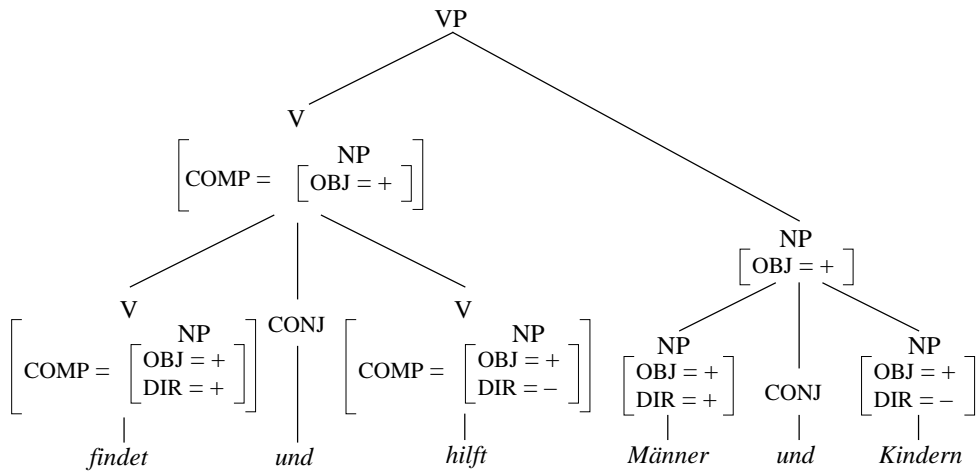

Figure 8: The feature structure subsumption analysis of the ungrammatical (6).

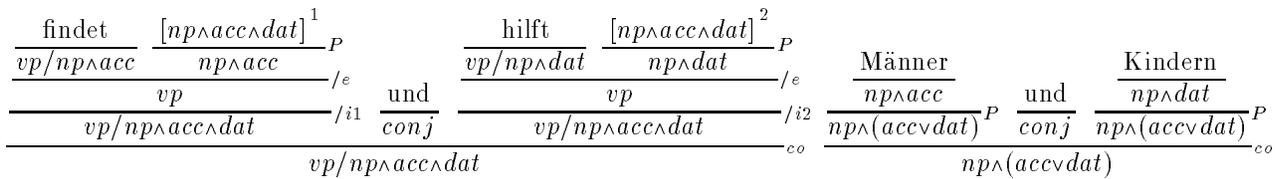

Figure 9: The blocked LCG analysis of the ungrammatical (6)

There are several approaches which may eliminate the need for such systematic ambiguity. First, if the language of (category) types is extended to permit *universally quantified* types as suggested by Morrill (Morrill, 1992), then adverbs could be assigned to the single type

$$\forall X.((s\backslash np \wedge X)\backslash(s\backslash np \wedge X)).$$

Second, it might be possible to reanalyse adjunction in such a way that avoids the problem altogether. For example, Bouma and van Noord (1994) show that assuming that heads subcategorize for adjuncts (rather than the other way around, as is standard) permits a particularly elegant account of the double infinitive construction in Dutch. If adjuncts in general are treated as arguments of the head, then the 'problem' of 'passing features' through adjunction disappears.

The comparative computational complexity of both the unification-based approach and the LCG accounts is also of interest. Despite their simplicity, the computational complexity of the kinds of feature-structure and LCG grammars discussed here is largely unknown. Dorre et. al. (1992) showed that the satisfiability problem for systems of feature-structure subsumption and equality constraints is undecidable, but it is not clear if such problems can arise in the kinds of feature-structure grammars discussed above. Conversely, while terminating (Gentzen) proof procedures are available for extended LCG systems of the kind we presented here, none of these handle the coordination schema, and as far as we are aware the computational properties of systems which include this schema are largely unexplored.